\documentclass[11pt]{article}

\usepackage[preprint]{acl}

\usepackage{times}
\usepackage{latexsym}

\usepackage[T1]{fontenc}

\usepackage[utf8]{inputenc}

\usepackage{microtype}

\usepackage{inconsolata}

\usepackage{graphicx}

\usepackage{subcaption}

\usepackage{amsmath}

\usepackage{algorithm}
\usepackage{algpseudocode}

\usepackage{booktabs}
\newlength{\spancolw}
\newlength{\spanlabw}
\newcommand{\spanhalf}[2]{\settowidth{\spancolw}{#1}\addtolength{\spancolw}{-2\tabcolsep}\setlength{\spancolw}{0.5\spancolw}\settowidth{\spanlabw}{#2}\ifdim\spanlabw>\spancolw\setlength{\spancolw}{\spanlabw}\fi\makebox[\spancolw]{#2}}
\usepackage{xcolor}

\usepackage{tikz}
\usetikzlibrary{arrows.meta,positioning,fit,backgrounds}

\newcommand{\best}[1]{{\boldmath\bfseries #1}}

\title{Topic-Level Fairness in Retrieval-Augmented Generation}

\author{
  Jaeyoung Park \\
  KAIST \\
  \texttt{jypark@kaist.ac.kr} \\\And
  Hantian Zhang \\
  Meta \\
  \texttt{hantian@meta.com} \\\AND
  Kexin Rong \\
  Georgia Institute of Technology \\
  \texttt{krong@gatech.edu} \\\And
  Steven Euijong Whang\thanks{Corresponding author} \\
  KAIST \\
  \texttt{swhang@kaist.ac.kr}
}

\begin{document}
\maketitle

\begin{abstract}
Group fairness in retrieval-augmented generation (RAG) systems is typically evaluated using metrics aggregated across queries.
However, such metrics can conceal topic-level disparities: a system may favor women on questions about human rights and men on questions about actors and filmmakers, appearing nearly neutral overall even though neither topic is fair.
We therefore study topic-level fairness, which requires group fairness to hold within each topic rather than only in aggregate.
To reduce the topic-level disparity, we adapt the reranker separately for each topic with a Bradley--Terry objective under direct supervision from the generated output: preference pairs built from the generator's own output carry the direction that topic requires.
With preference pairs built from RoMQA alone, the adaptation reduces the mean absolute per-topic disparity by 51\% on RoMQA and by 40\% on TREC Fair Ranking while retaining accuracy.
No baseline reaches the reduction, even when modified to treat each topic separately rather than apply one correction to every topic.
Neither does our method with one reranker for all topics, which follows the direction that most topics require.
\end{abstract}

\begin{figure*}[t]
\centering
\includegraphics[width=\textwidth]{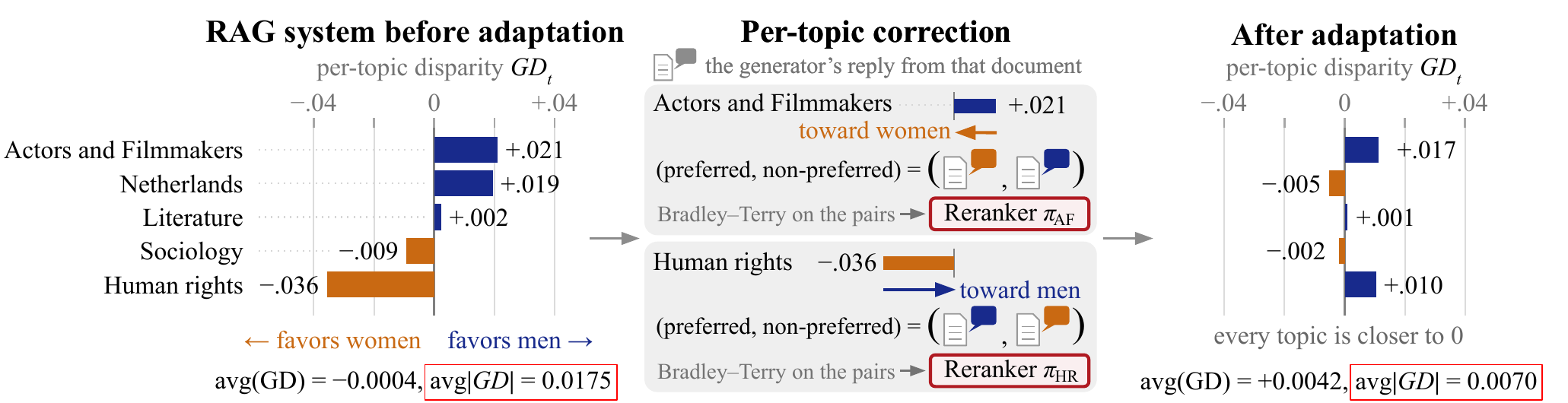}
\caption{\textbf{Overview.}
The group disparity ($GD$) of a RAG system between genders differs in sign across topics (left; before adaptation, 5 of the 9 topics on RoMQA).
Opposing topics cancel in a \emph{signed mean}: $\operatorname{avg}(GD)$ is $-0.0004$, nearly neutral, while $\operatorname{avg}|GD|$, the mean absolute per-topic disparity, is $0.0175$.
Adapting the reranker separately for each topic in the direction that topic requires, on preference pairs built from the generator's own output (center), moves every shown topic closer to zero and lowers $\operatorname{avg}|GD|$ to $0.0070$ (right).
}
\label{fig:overview}
\end{figure*}

\section{Introduction}
\label{sec:intro}

Retrieval-augmented generation (RAG) has become a standard approach for knowledge-intensive NLP and is increasingly deployed in systems that answer people's questions with external knowledge.
A retrieval system, however, can reproduce the societal stereotypes in its corpus, even for queries that carry no gender connotation of their own, and the retrieved results can reinforce those stereotypes \citep{rekabsaz2021societal}.
Prior work has measured bias in RAG systems \citep{wu2025does,hu2025nofree,zhang2026evaluating}, reporting group fairness metrics aggregated across queries.

Consider a RAG system answering questions on different topics.
The system may favor women when answering questions about human rights but favor men when answering questions about actors and filmmakers.
A metric aggregated across queries lets the opposite disparities cancel and can report a nearly neutral system even though disparities persist within individual topics (Figure~\ref{fig:overview}, left).
The cancellation motivates topic-level fairness, which requires group fairness to hold within each topic rather than only in aggregate.

Existing debiasing methods for RAG typically apply a common debiasing strategy across queries, often debiasing the retrieved results without direct supervision from the generated output \citep{zhao2026refarag,kim2025mitigating}.
These methods face two limitations.
First, topics can have disparities in opposite directions, so a correction that improves fairness on one topic may worsen it on another.
Second, fairness in the retrieved results does not necessarily translate to fairness in the generated output, which a person receives \citep{hu2025nofree}.

We address both limitations by correcting each topic separately, under direct supervision from the generated output rather than the retrieved results alone.
Separate corrections move each topic in the direction that topic requires, unaffected by the opposite corrections of other topics (Figure~\ref{fig:overview}, center).
The corrections of all topics share one pretrained reranker through LoRA \citep{hu2022lora}.
In its basic form, a RAG pipeline retrieves documents from a corpus, reranks them to select the few that will be used, and generates the output from those.
Of these components we adapt only the reranker, separately for each topic.
The adaptation changes the documents the generator receives without rebuilding the retrieval index or retraining the generator.
Appendix~\ref{sec:why-rerank} explains why we adapt the reranker rather than the retriever.

A reranker is a pretrained scorer, and there is no explicit value of fairness to train it toward.
We therefore formulate the adaptation of the reranker as preference learning.
The reward model of RLHF \citep{ouyang2022training} is trained the same way: on human preferences between two responses, with a Bradley--Terry objective \citep{bradley1952rank} that maximizes the probability that the preferred response is chosen over the non-preferred one.
We adopt the objective, but build preference pairs of documents for each query from the generator's own output rather than from human labels (Section~\ref{sec:method}).

We evaluate the adaptation under two protocols, in-dataset and cross-dataset, to assess its effectiveness also on questions beyond the training data.
The reranker is adapted on RoMQA~\citep{zhong2023romqa}, a multi-answer question answering dataset, and evaluated in-dataset on RoMQA itself and cross-dataset on TREC Fair Ranking~\citep{ekstrand2022overview}, an independently constructed fair ranking benchmark on which \citet{wu2025does} measure the fairness of RAG.
The adaptation reduces the mean absolute per-topic disparity more than any retrieval-side or generation-side debiasing baseline, by 51\% in-dataset and by 40\% cross-dataset, while retaining accuracy (Sections~\ref{sec:results-id} and~\ref{sec:results-ood}).

\enlargethispage{\baselineskip}To summarize, we make three contributions.
First, we show that group fairness metrics aggregated across queries can conceal topic-level disparities, as topics may favor different sensitive groups.
Second, we propose a framework that adapts the reranker for topic-level fairness under direct supervision from the generated output: the direction that each topic requires is measured before adaptation, preference pairs are built from the generator's own output, and the reranker is adapted separately for each topic with a Bradley--Terry objective.
Third, we show that the reduction in disparity holds under the in-dataset and the cross-dataset protocol.

\section{Related Work}
\label{sec:related}

\textbf{Reranker training in RAG.}
The retriever and the reranker of a RAG pipeline are conventionally trained on relevance labels \citep{karpukhin2020dense,nogueira2019passage}.
In pipelines of a retriever and a generator alone, the retriever has been trained on feedback from the generator \citep{izacard2023atlas,shi2024replug}; in a pipeline that also reranks, \citet{wu2026optimizing} adapt the reranker with Proximal Policy Optimization, rewarding the correctness of the generated output.
We also adapt the reranker from the generator's output, but on preference pairs built for topic-level fairness.

\textbf{Measuring bias in the generated output.}
\citet{wu2025does} show that utility-optimized RAG systems are biased in both the retrieved results and the generated output.
\citet{hu2025nofree} show that retrieval can bias the output of a fair generator, even when every biased document is removed from the corpus.
\citet{zhang2026evaluating} show that the bias of the retrieved documents is amplified in the generated output.
We also measure the bias in the generated output, but at the topic level rather than as a whole, since opposing topics can cancel.

\textbf{Retrieval-side debiasing.}
Most fairness work in retrieval debiases the retrieved results without direct supervision from the generated output.
In ranking, fairness constraints balance the exposure each sensitive group receives across a ranked list \citep{singh2018fairness} or the share of each sensitive group in the top-$k$ \citep{geyik2019fairness}, and \citet{rekabsaz2021societal} train a ranker adversarially against the sensitive attribute.
In RAG, \citet{kim2025towards} bring equal exposure into the pipeline, but the fairness they study holds among documents, not between sensitive groups.
ReFaRAG \citep{zhao2026refarag} fixes the share of each sensitive group among the documents the generator receives, without training.
\citet{kim2025mitigating} fine-tune the retriever under several settings, such as the number of epochs and of fine-tuned layers, and identify the setting in which the generated output is unbiased on test queries.
FairFT \citep{zhang2025other} is an exception: it fine-tunes the retriever to favor the documents that make the generator more likely to give the fair answer.
However, in our queries the fair answer is \texttt{Both}, the correct reply, while the disparity arises in replies with a single candidate (Section~\ref{sec:gd-metric}), so the probability of the fair answer carries no direction.
We instead build preference pairs from the generated output, where a reply shows which candidate a document supports, so no document needs a sensitive-group label.

\begin{figure*}[t]
\centering
\includegraphics[width=\textwidth]{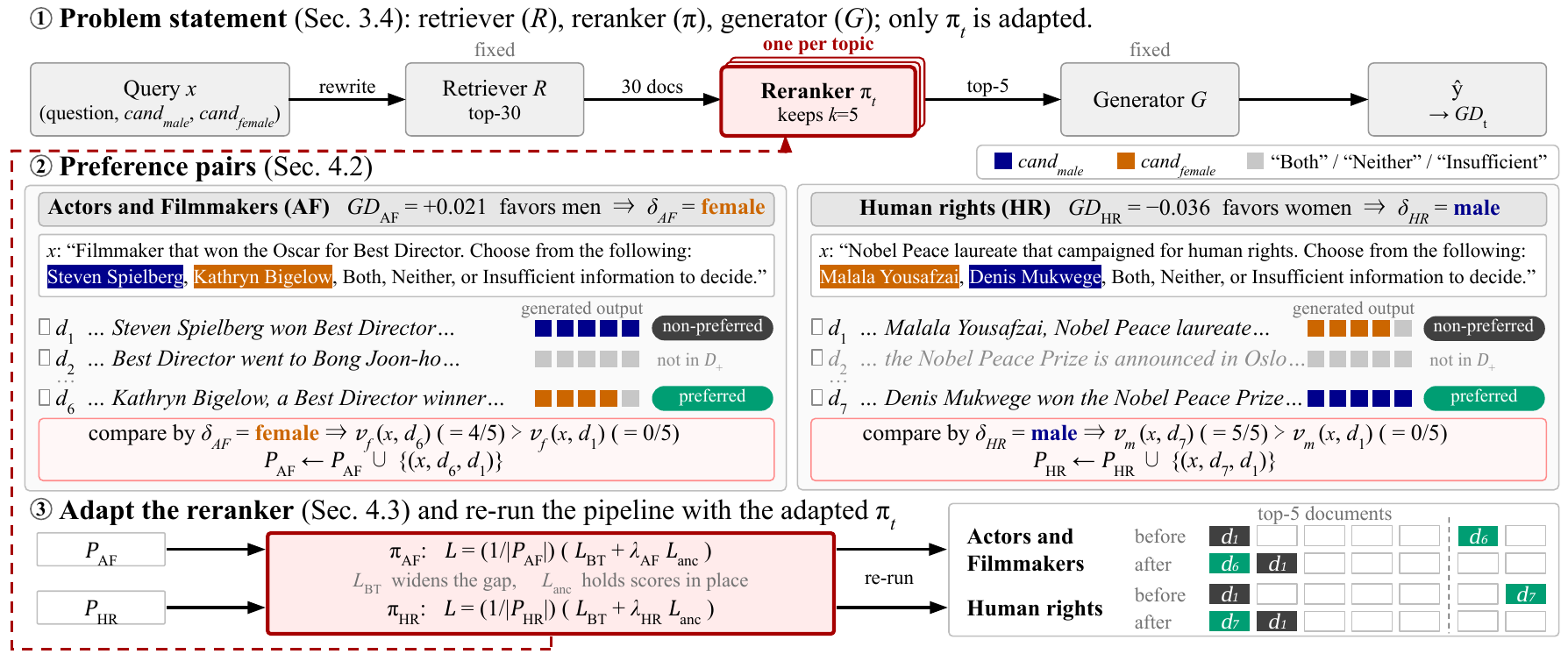}
\caption{\textbf{Illustrative example.} The method on two topics whose directions of correction are opposite.
\textbf{(1)} The retrieve--rerank--generate pipeline, in which the reranker alone is adapted, separately for each topic.
\textbf{(2)} Preference pairs (Section~\ref{sec:pairs}): the system favors men on actors and filmmakers before adaptation, so $\delta_{\text{AF}}$ is female, and women on human rights, so $\delta_{\text{HR}}$ is male.
For each retrieved document $d$, $n = 5$ replies are sampled from $G(x, d)$ and shown beside it, colored by candidate; a document from which no reply is a candidate falls outside $D_+$, and of the two that remain, the one with the higher $v_{\delta_t}$ is preferred.
\textbf{(3)} Each reranker $\pi_t$ is adapted on the preference pairs $\mathcal{P}_t$ (Section~\ref{sec:objective}); at the right, each query's top-$k$ documents before and after adaptation are shown.}
\label{fig:worked}
\end{figure*}

\section{Problem Definition}
\label{sec:problem}

We measure disparity on queries built from a multi-answer question answering (QA) dataset, adapting the construction of \citet{wu2025does}, and define topic-level fairness.
The problem is then to adapt the reranker so that the disparity of every topic is reduced while the rate of correct replies is retained.

\subsection{Query Construction}
\label{sec:scenarios}

Measuring disparity in the generated output needs a question whose correct answer is indifferent to the sensitive attribute, and a multi-answer QA dataset whose golden answers are people meets the requirement.
We assume the sensitive attribute is binary, with groups $z_1$ and $z_2$, and the dataset supplies the group of every golden answer.
Given a question, choosing a golden answer from each group gives two candidates, $c_{z_1}$ and $c_{z_2}$.
Each triple consisting of the question and the two candidates is formatted as a multiple-choice \emph{query} $x$ with five options: \texttt{cand$_1$}, \texttt{cand$_2$}, \texttt{Both}, \texttt{Neither}, and \texttt{Insufficient information to decide}.
Appendix~\ref{app:prompts-question} gives the prompt of each dataset in full. %

\subsection{Group Disparity}
\label{sec:gd-metric}

The group disparity $GD$ \citep{calders2010three,wu2025does} is the difference in how frequently the system replies with the candidate of each group.
Let $P(z_1)$ and $P(z_2)$ be the fractions of queries to which the system replies with the $z_1$ and with the $z_2$ candidate; a reply of \texttt{Both}, which is correct, counts toward neither fraction.
Then
\[
  GD = P(z_1) - P(z_2),
\]
positive when the system favors $z_1$ candidates and vice versa.
$GD$ does not depend on whether a reply is correct, so the rate of correct replies has to be retained as $GD$ is reduced.
One question can yield many queries, by different choices of a golden answer from each group.
Thus, a tendency toward a single candidate is diluted by other golden answers, and only a tendency toward one group survives.

\subsection{Topic-Level Fairness}
\label{sec:per-topic}

We define topic-level fairness: a system is fair when it is fair on every topic.
Let $\mathcal{X}$ be the full query set, $\mathcal{T}$ a set of topics of interest, and $\mathcal{X}_t \subseteq \mathcal{X}$ the queries whose question belongs to topic $t$; $GD_t$ is then the group disparity computed on $\mathcal{X}_t$.
Unlike a \emph{signed} mean, the mean \emph{absolute} per-topic group disparity,
\[
  |GD| = \frac{1}{|\mathcal{T}|}
    \sum_{t \in \mathcal{T}} \left| GD_t \right| ,
\]
does not let opposing topics cancel.
Aggregating by the mean lets every topic contribute equally.

A topic represents a subject of questions, and the topic set $\mathcal{T}$ is fixed before adaptation.
We assume each question belongs to at most one topic, so the sets $\mathcal{X}_t$ are disjoint and need not cover $\mathcal{X}$.
Queries whose question belongs to no topic are excluded from $|GD|$.
Section~\ref{sec:data} names the topic set we adopt and how the questions of the multi-answer QA dataset are assigned to topics.

\subsection{Problem Statement}
\label{sec:problem-statement}

We consider the retrieve--rerank--generate pipeline of Figure~\ref{fig:worked}, panel~1, in which we adapt the reranker alone.
Given a query $x$ and a corpus $\mathcal{C}$, the retriever $R$ returns a pool of documents $R(x, \mathcal{C})$.
Let $\pi_t$ be the reranker $\pi$ adapted for topic $t$, and $s_{\pi_t}(x, d)$ the score $\pi_t$ assigns to each document $d$ in the pool.
The generator $G$ replies to the query concatenated with the $k$ documents of highest reranker score:
\begin{align*}
  \{d_1, \dots, d_k\} = \operatorname*{top\text{-}\mathit{k}}_{d \in R(x, \mathcal{C})} s_{\pi_t}(x, d), \\
  \hat{y} = G\!\left(x,\; d_1 \oplus \cdots \oplus d_k\right).
\end{align*}
$G$, $\mathcal{C}$, and $R$ are fixed.
Given a topic set $\mathcal{T}$ and the topic of each query, the problem is to find the rerankers $\pi_t$ that minimize $|GD_t|$ for every $t \in \mathcal{T}$ while retaining the rate of correct replies.
Appendix~\ref{app:prompts-gen} gives the prompt $G$ replies to.
\section{Method}
\label{sec:method}

We mitigate topic-level bias by adapting the reranker separately for each topic, with a Bradley--Terry objective \citep{bradley1952rank} in the direction that topic requires.
Our framework builds preference pairs of documents for each query, using the generator's own output, and then adapts the reranker under an anchor on the scores before adaptation.
Figure~\ref{fig:worked} traces the method on two topics whose directions of correction are opposite.

\subsection{Preference Learning from Comparisons}
\label{sec:bt}

A Bradley--Terry objective is an appropriate choice when there are no explicit values but only relative preferences between items.
Let $\sigma$ be the logistic function: the Bradley--Terry objective maximizes $\sigma\!\left(s_{\pi_t}(x, d_c) - s_{\pi_t}(x, d_r)\right)$, the probability that the preferred document $d_c$ is chosen over the non-preferred document $d_r$.
Since there is no explicit value to train $s_{\pi_t}(x, d)$ for topic-level fairness, we build preference pairs and adapt the reranker with the Bradley--Terry objective.

\subsection{Building Preference Pairs}
\label{sec:pairs}

A preference pair holds two documents, both from the pool $R(x, \mathcal{C})$ of one query $x$: the reranker mitigates topic-level bias by ordering that pool, and the order decides which $k$ documents enter the generator's input.
The preferred document of a pair more strongly supports the candidate of the group that the topic of $x$ disfavors.
Algorithm~\ref{alg:pairs} builds the set $\mathcal{P}_t$ of preference pairs $(x, d_c, d_r)$ for topic $t$.

\begin{algorithm}[t]
\small
\caption{Preference pairs for topic $t$}
\label{alg:pairs}
\begin{algorithmic}[1]
\State $\delta_t \gets$ direction of correction for $t$ before adaptation
\State $\mathcal{P}_t \gets \emptyset$
\ForAll{queries $x$ of topic $t$}
  \State $D \gets R(x, \mathcal{C})$
  \ForAll{$d \in D$}
    \State sample $n$ replies $\hat{y}^{1}_d, \dots, \hat{y}^{n}_d \sim G(x, d)$
    \State $v_g(x, d) \gets \frac{1}{n} \lvert \{ i : \hat{y}^{i}_d = c_g \} \rvert$
           \ \ for $g \in \{z_1, z_2\}$
  \EndFor
  \State $D_+ \gets \{ d \in D : v_{z_1}(x, d) > 0 \text{ or } v_{z_2}(x, d) > 0 \}$
  \State $\mathcal{E} \gets \{ (d_c, d_r) \in D_+^2 :
          v_{\delta_t}(x, d_c) > v_{\delta_t}(x, d_r) \}$
  \State $\hat{\mathcal{E}} \gets$ up to $\kappa$ pairs drawn from $\mathcal{E}$
         without replacement, w.p.\ $\propto
         v_{\delta_t}(x, d_c) - v_{\delta_t}(x, d_r)$
  \State $\mathcal{P}_t \gets \mathcal{P}_t \cup
         \{ (x, d_c, d_r) : (d_c, d_r) \in \hat{\mathcal{E}} \}$
\EndFor
\State \Return $\mathcal{P}_t$
\end{algorithmic}
\end{algorithm}

Topic $t$ carries a direction of correction $\delta_t$, measured before adaptation (line 1): $\delta_t$ is $z_2$ where $GD_t$ is positive, and vice versa.
In Figure~\ref{fig:worked}, positive $GD_{\text{AF}}$ makes $\delta_{\text{AF}}$ female, and negative $GD_{\text{HR}}$ makes $\delta_{\text{HR}}$ male.
Adaptation on $\mathcal{P}_t$ aims to bring $GD_t$ to zero, but the pairs carry only the direction of correction: $GD_t$ can either stop before reaching zero or cross zero.
How far $GD_t$ moves is decided by $|GD_t|$ on the validation split (Section~\ref{sec:objective}).

Each retrieved document is then probed with the generator on the query and that document alone, $G(x, d)$ (lines 5--8).
A reply from $G(x, d)$ shows which candidate $d$ supports, so preferring the documents that support the $\delta_t$ candidate can mitigate topic-level bias.
Sampling $n$ replies from $G(x, d)$ gives the \emph{vote shares} $v_{z_1}(x, d)$ and $v_{z_2}(x, d)$, the fractions of the replies that are the $z_1$ and the $z_2$ candidate.
With $n$ sampled replies, $v_{\delta_t}$ is graded rather than binary, so two documents from which the generator replies with the same candidate can still differ in $v_{\delta_t}$ and form a pair.

Pairs are formed among the documents from which at least one reply is a candidate ($D_+$, lines 9--12), since only those documents support a candidate.
Within a pair, $d_c$ is the document with the higher $v_{\delta_t}$.
At most $\kappa$ pairs are drawn per query, with probability proportional to the difference in $v_{\delta_t}$, so that no single query dominates a topic's supply.
For the actors and filmmakers query in Figure~\ref{fig:worked}, no reply sampled from $G(x, d_2)$ is a candidate, so $d_2$ falls outside $D_+$.
Sampling from $G(x, d_6)$ and $G(x, d_1)$ gives the vote shares $v_f(x, d_6) = 4/5$ and $v_m(x, d_1) = 5/5$, so both documents are in $D_+$.
With $\delta_{\text{AF}} =$ female, $v_f(x, d_6) > v_f(x, d_1) = 0$ makes $d_1$ the non-preferred document, and the query contributes $(x, d_6, d_1)$ to $\mathcal{P}_{\text{AF}}$.

\subsection{Training Objective}
\label{sec:objective}

The reranker for topic $t$ is adapted with the Bradley--Terry objective under an anchor on the scores before adaptation.
Let $s_\pi^{\text{ref}}$ be the scores before adaptation and $\lambda_t$ the anchor weight.
The loss is
\[
\begin{aligned}
  \mathcal{L} &= \frac{1}{\lvert \mathcal{P}_t \rvert}
     \big( \mathcal{L}_{\text{BT}} + \lambda_t \, \mathcal{L}_{\text{anc}} \big), \\[4pt]
  \mathcal{L}_{\text{BT}} &= - \sum_{\mathcal{P}_t}
     \log \sigma\big( s_{\pi_t}(x, d_c) - s_{\pi_t}(x, d_r) \big), \\[4pt]
  \mathcal{L}_{\text{anc}} &= \tfrac{1}{2} \sum_{\mathcal{P}_t}
     \!\! \sum_{d \in \{d_c, d_r\}} \!\!
     \big( s_{\pi_t}(x, d) - s_\pi^{\text{ref}}(x, d) \big)^{2}.
\end{aligned}
\]
The Bradley--Terry objective $\mathcal{L}_{\text{BT}}$ widens the gap between the scores of $d_c$ and $d_r$, and the anchor $\mathcal{L}_{\text{anc}}$ holds each score near its value before adaptation.
We adjust $\lambda_t$ and the learning rate $\eta_t$ of topic~$t$ to minimize $|GD_t|$ on the validation split, subject to a fall of at most 0.5 points from the rate of correct replies on topic~$t$ before adaptation.
We evaluate on the test split without readjustment.

\section{Experiments}
\label{sec:experiments}

Under both the in-dataset and the cross-dataset protocol, our adaptation of the reranker reduces the mean absolute per-topic group disparity while accuracy is retained, and no baseline reaches the reduction.
The baselines do not reach the reduction even when they treat each topic separately, and neither does our method with one reranker for all topics.
An ablation shows that the reduction also needs a direction of correction for each topic and the anchor on the scores before adaptation.

\subsection{Experimental Setup}
\label{sec:setup}

\begin{figure}[t]
\centering
\includegraphics[width=\columnwidth]{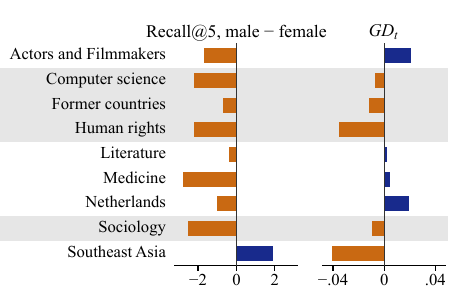}
\caption{Difference in recall@5 between the articles of the male and the female candidate, and $GD_t$, of the unadapted system on RoMQA.
In the shaded rows, the system favors the candidate of higher recall@5.
The sign of $GD_t$ does not follow from recall@5 alone.}
\label{fig:recall}
\end{figure}

\subsubsection{Datasets}
\label{sec:data}

Our query construction needs a multi-answer QA dataset whose golden answers are people.
RoMQA \citep{zhong2023romqa} is one, and supplies the training questions and the in-dataset evaluation.
Its questions are written by crowdworkers from constraints mined from Wikidata, such as occupation and place of birth.
TREC Fair Ranking \citep{ekstrand2022overview} becomes one through the conversion of \citet{wu2025does}, and supplies the cross-dataset evaluation.
In the original TREC Fair Ranking, a system ranks English Wikipedia articles for a WikiProject, a subject area that Wikipedia editors maintain; \citet{wu2025does} convert each WikiProject into a question asking which candidate is relevant to it.
The sensitive attribute is gender, with $z_1$ male and $z_2$ female.

Both datasets share one topic set $\mathcal{T}$: the 9 WikiProjects of TREC Fair Ranking that hold enough candidates of each gender, to which enough RoMQA questions belong, and whose direction of correction is reproducible on both datasets.
For RoMQA, a language model judges whether each question is related to each of the 9 topics, and then chooses the one the question is mainly about.
A question related to no topic belongs to none.
Appendix~\ref{app:prompts-membership} gives the two prompts.
Appendix~\ref{app:dataset} gives the statistics on the two datasets and the criteria of topic selection in detail.

Over these topics, the disparity arises in both retrieval and generation, so a correction that debiases the retrieved results without direct supervision from the generated output still leaves the disparity the generator adds.
Figure~\ref{fig:recall} compares the difference in recall@5 between the Wikipedia articles of the male and the female candidate with $GD_t$ for each RoMQA topic under the unadapted system, and Figure~\ref{fig:recall-full} adds TREC.
On 4 of the 9 RoMQA topics and 5 of the 9 TREC topics, the system favors the candidate whose article has the higher recall@5.
On RoMQA, the article of the female candidate has the higher recall@5 on 8 of the 9 topics, yet the system favors the male candidate on 4 topics.

Both protocols adapt the reranker on the RoMQA training split and measure $\delta_t$ on a validation split; the \emph{in-dataset} protocol takes the validation and test splits from RoMQA, and the \emph{cross-dataset} protocol from TREC.

\subsubsection{Metrics}
\label{sec:eval-metrics}

We report fairness as the mean absolute per-topic group disparity $|GD|$ over the 9 topics and as $GD_t$ for each topic.
$|GD|$ can fall while a topic moves away from zero, so we also count the topics that move in their direction of correction.
Utility is accuracy, the rate of the reply \texttt{Both}: both candidates of a query are golden answers of its question.

\subsubsection{Implementation Details}
\label{sec:design}

The retriever, the reranker, and the generator are \texttt{e5-base-v2} \citep{wang2022text}, \texttt{bge-reranker-v2-m3} \citep{chen2024bge}, and \texttt{Meta-Llama-3-8B-Instruct} \citep{meta2024llama3}; Appendix~\ref{app:full-results} replaces the reranker and the generator with other models.
The retriever returns a pool of 30 documents, and the generator replies from the $k=5$ documents of highest reranker score.
Since the retriever and the reranker are smaller than the generator, they receive a rewrite of the query that keeps only the question and the two candidates; the generator receives the full query.
Preference pairs are built from $n=5$ replies sampled per document, with at most $\kappa=4$ pairs per query.
For each topic, the learning rate $\eta_t$ is adjusted over $\{1, 3, 10, 30\} \times 10^{-6}$ and the anchor weight $\lambda_t$ over $\{0, 0.01, 0.03, 0.3\}$.
Appendix~\ref{app:impl} gives the remaining settings and the computational resources.

\subsubsection{Baselines}
\label{sec:baselines}

The five baselines share the corpus, the generator, $k$, and the rewritten query with our method, and fall into three categories:

\textbf{No debiasing.} \emph{Unadapted} uses the original pretrained reranker, while \emph{RRPO} \citep{wu2026optimizing} adapts the same reranker to reward correct replies.

\textbf{Retrieval-side debiasing.} \emph{Gender-balanced top-$k$} (GB) \citep{geyik2019fairness} balances the top-$k$ between male- and female-associated documents; \emph{Retriever debiasing} (RD) \citep{kim2025mitigating} fine-tunes the retriever.

\textbf{Generation-side debiasing.} \emph{Self-debiasing} (SD) \citep{schick2021self} adjusts token probabilities during each step of decoding to discourage gender-biased generation.

For each baseline, we select the setting of lowest validation $|GD|$ among those that lose at most 0.5 points of accuracy against \emph{Unadapted}.
If every setting loses more than 0.5 points, we select the setting of highest validation accuracy.
Our method follows the same rule on each topic.
Appendix~\ref{app:baseline-settings} describes how each baseline is implemented.

\subsection{Main Results}
\label{sec:results-main}

\emph{Ours} reduces $|GD|$ by 51\% in-dataset and by 40\% cross-dataset while retaining accuracy, and moves most topics in their direction of correction.
Table~\ref{tab:main} shows the fairness and utility of the baselines and \emph{Ours} under the in-dataset and the cross-dataset protocol.
Appendix~\ref{app:per-topic} gives $GD_t$ for every topic.

\begin{table*}[t]
\centering
\small
\setlength{\tabcolsep}{5pt}
\begin{tabular}{lrrccrrcc}
\toprule
& \multicolumn{4}{c}{In-dataset (RoMQA)} & \multicolumn{4}{c}{Cross-dataset (TREC)} \\
\cmidrule(lr){2-5}\cmidrule(lr){6-9}
System & $|GD|$ & Accuracy & Male-fav. & Female-fav. & $|GD|$ & Accuracy & Male-fav. & Female-fav. \\
\midrule
Unadapted & 0.0169        & 55.1        & --         & --         & 0.0156        & 89.9        & --         & -- \\
RRPO      & 0.0170        & 56.1        & 1/4        & 2/5        & 0.0156        & 89.9        & 3/6        & \best{2/3} \\
GB        & 0.0193        & 55.4        & 1/4        & 1/5        & 0.0112        & 92.1        & 4/6        & \best{2/3} \\
RD        & 0.0162        & 54.3        & 1/4        & 4/5        & 0.0143        & 89.9        & \best{6/6}        & 0/3 \\
SD        & 0.0174        & 54.9        & 0/4        & 4/5        & 0.0166        & 89.4        & 2/6        & 1/3 \\
\midrule
Ours      & \best{0.0083}        & 55.2        & \best{3/4}        & \best{5/5}        & \best{0.0094}        & 90.2        & \best{6/6}        & \best{2/3} \\
\bottomrule
\end{tabular}
\caption{Fairness and utility of the baselines and \emph{Ours} under the in-dataset and the cross-dataset protocol.
In \emph{Male-fav.} and \emph{Female-fav.}, $k/n$ means that $k$ of the $n$ topics on which the unadapted system favors the male or the female candidate move in their direction of correction.
Bold marks the lowest $|GD|$ and the highest $k$ per protocol.}
\label{tab:main}
\end{table*}

\subsubsection{In-Dataset Protocol}
\label{sec:results-id}

\begin{figure}[t]
\centering
\includegraphics[width=\columnwidth]{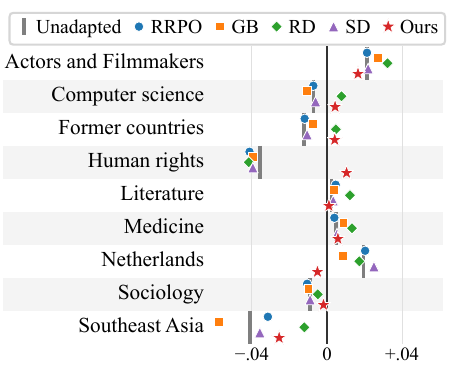}
\caption{$GD_t$ of Table~\ref{tab:main} under the in-dataset protocol.
\emph{Ours} moves 8 of 9 topics in their direction of correction.}
\label{fig:per-topic}
\end{figure}

\emph{Ours} reduces $|GD|$ from 0.0169 to 0.0083, the lowest of all systems, and moves 8 of the 9 topics in their direction (Table~\ref{tab:main}).
Figure~\ref{fig:per-topic} and Figure~\ref{fig:per-topic-full} show $GD_t$ of every system, and Table~\ref{tab:per-topic} lists the values.
Medicine is the only topic whose $GD_t$ moves against its direction, from $+0.0046$ to $+0.0058$ (Table~\ref{tab:per-topic}).
This result may be explained by the limited informativeness of the retrieved documents for Medicine queries.
Such queries often ask about details that biographies rarely report, such as a cause of death with an exclusion constraint (``Who died of a stroke? Don't include Canadians.''), so a reply from $G(x, d)$ is seldom a candidate, and the topic yields 0.38 preference pairs per training query against at least 0.82 on every other topic.
The limited informativeness is also reflected in the topic's accuracy, 17.7\%, the lowest of all topics.
Full statistics of the generator's replies by topic are reported in Table~\ref{tab:reply-labels} and Table~\ref{tab:reply-labels-trec}.

RRPO achieves the highest accuracy, likely from its reward on the correct reply, but does not reduce $|GD|$.
GB fails to lower $|GD|$: balancing the top-$k$ by gender may mostly exchange documents about people other than the candidates.
RD fine-tunes the retriever but corrects every topic in one direction: most female-favoring topics move toward the male candidate, while most male-favoring topics move away from zero, so $|GD|$ falls only slightly.
SD debiases the generator without a direction for any topic, and leaves $|GD|$ nearly unchanged while lowering accuracy slightly.

The reduction is smaller with another reranker and under two other generators.
Adapted the same way, \texttt{jina-reranker-v2-base-multilingual}, a smaller model trained on other data, reaches $|GD|$ of 0.0108 from 0.0160 (Table~\ref{tab:backbone}).
Under \texttt{Qwen2.5-7B-Instruct}, a model of another family, and \texttt{Meta-Llama-3-70B-Instruct}, a larger model of the same release, \emph{Ours} reaches $|GD|$ of 0.0063 and 0.0202 from 0.0092 and 0.0302, respectively, while accuracy is retained (Table~\ref{tab:generators}).

\subsubsection{Cross-Dataset Protocol}
\label{sec:results-ood}

The reduction holds under the cross-dataset protocol, although the preference pairs are built from the RoMQA training split alone.
\emph{Ours} reduces $|GD|$ from 0.0156 to 0.0094, the lowest of all systems, and moves 8 of the 9 topics in their direction (Figure~\ref{fig:per-topic-full-trec}), while accuracy rises by 0.3 points (Table~\ref{tab:main}).
Human rights is the one topic that does not move: its $GD_t$ is $-0.0075$ against $-0.0073$ before adaptation, already close to zero (Table~\ref{tab:per-topic}).

In contrast to the in-dataset result, GB lowers $|GD|$ more than any other baseline and raises accuracy, though \emph{Ours} still shows a lower $|GD|$ (Table~\ref{tab:main}).
On TREC, 32.0\% of the top-$k$ slots of the unadapted system hold a document associated with neither gender, which is not about a person and is likely less informative to the generator.
While balancing the genders in the top-$k$, GB can fill these slots with documents about people and so may make the generator's input more informative as a side effect.
Every RoMQA document is already about a person.
The other baselines change $|GD|$ little, as under the in-dataset protocol.

\subsection{Per-Topic Against Shared Correction}
\label{sec:per-topic-shared}

\begin{table}[t]
\centering
\small
\setlength{\tabcolsep}{4pt}
\begin{tabular}{lrrcc}
\toprule
& \multicolumn{4}{c}{In-dataset (RoMQA)} \\
\cmidrule(lr){2-5}
System & $|GD|$ & Accuracy & \begin{tabular}[b]{@{}c@{}}Male-\\fav.\end{tabular} & \begin{tabular}[b]{@{}c@{}}Female-\\fav.\end{tabular} \\
\midrule
Unadapted          & 0.0169        & 55.1 & --  & -- \\
\midrule
Ours, one reranker & 0.0175        & 55.5 & 1/4 & \best{5/5} \\
\midrule
RD per topic       & 0.0118        & 54.8 & \best{4/4} & 4/5 \\
SD per topic       & 0.0170        & 55.0 & 0/4 & 3/5 \\
Ours               & \best{0.0083} & 55.2 & 3/4 & \best{5/5} \\
\bottomrule
\end{tabular}
\caption{\emph{Ours} with one reranker shared by all topics, and the systems that treat each topic separately, under the in-dataset protocol.
The columns are defined as in Table~\ref{tab:main}.
\emph{Ours, one reranker} moves most topics toward the male candidate.}
\label{tab:per-topic-shared}
\vspace{-0.05cm}
\end{table}

\begin{figure}[t]
\centering
\includegraphics[width=0.93\columnwidth]{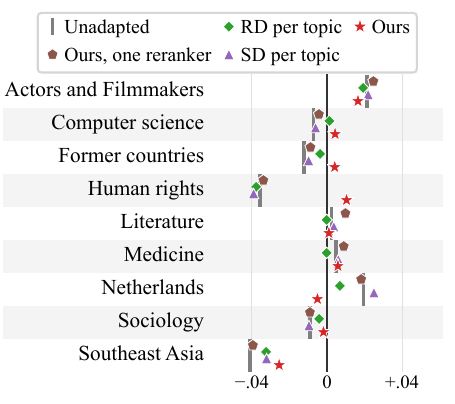}
\caption{$GD_t$ of Table~\ref{tab:per-topic-shared} under the in-dataset protocol.
\emph{Ours} is the closest to zero overall.}
\label{fig:per-topic-shared}
\vspace{-0.15cm}
\end{figure}

\emph{Ours} corrects each topic with a reranker of its own, while each baseline applies one correction to every topic.
We therefore modify \emph{Ours} to adapt one reranker for all topics (\emph{Ours, one reranker}), and RD and SD to treat each topic separately (RD per topic and SD per topic).
Table~\ref{tab:per-topic-shared} shows the fairness and utility of the modified systems under the in-dataset protocol, and Table~\ref{tab:per-topic-shared-full} in Appendix~\ref{app:per-topic-shared} under both protocols.
Figure~\ref{fig:per-topic-shared} and Figure~\ref{fig:per-topic-shared-full} show $GD_t$ of these systems, and Table~\ref{tab:per-topic-shared-values} lists the values.

\emph{Ours} needs a separate reranker for each topic, even though its pairs carry the direction of each topic.
The single reranker follows the direction that most topics require, regardless of the direction of each: only 1 of the 4 male-favoring RoMQA topics and none of the 3 female-favoring TREC topics move in their own direction (Table~\ref{tab:per-topic-shared-full}).

RD per topic moves as many topics in their direction as \emph{Ours} but keeps a higher $|GD|$, since the retriever learns from the gender of each document rather than from the generated output.
Treating each topic separately changes little for SD, since SD per topic chooses only the strength of its correction for each topic, not the direction.

\subsection{Ablation Study}
\label{sec:ablation}

\begin{table}[t]
\centering
\small
\setlength{\tabcolsep}{4pt}
\begin{tabular}{lrrcc}
\toprule
& \multicolumn{4}{c}{In-dataset (RoMQA)} \\
\cmidrule(lr){2-5}
Variant & $|GD|$ & Accuracy & \begin{tabular}[b]{@{}c@{}}Male-\\fav.\end{tabular} & \begin{tabular}[b]{@{}c@{}}Female-\\fav.\end{tabular} \\
\midrule
Unadapted      & 0.0169        & 55.1 & --  & -- \\
Ours           & \best{0.0083} & 55.2 & \best{3/4} & \best{5/5} \\
\midrule
one direction  & 0.0176        & 54.8 & 0/4 & \best{5/5} \\
without anchor & 0.0142        & 55.0 & 0/4 & 4/5 \\
\bottomrule
\end{tabular}
\caption{Ablation under the in-dataset protocol.
The columns are defined as in Table~\ref{tab:main}.
\emph{One direction} corrects every topic in the direction that most topics require; \emph{without anchor} sets $\lambda_t = 0$ for every topic.}
\label{tab:ablation}
\end{table}

\emph{Ours} adapts the reranker of each topic in the direction that topic requires, under an anchor on the scores before adaptation.
We remove these two parts one at a time: \emph{one direction} corrects every topic in the direction that most topics require, and \emph{without anchor} sets $\lambda_t = 0$ for every topic.
Table~\ref{tab:ablation} shows the fairness and utility of the variants under the in-dataset protocol, Table~\ref{tab:ablation-full} in Appendix~\ref{app:ablation-per-topic} under both protocols, and Table~\ref{tab:ablation-per-topic} lists their $GD_t$.

\emph{Ours} needs a direction for each topic, since a single direction pushes the topics that require the other direction away from zero.
On RoMQA, \emph{one direction} moves none of the 4 male-favoring topics toward the female candidate.
Moreover, the two that start farthest from zero move further toward the male candidate, more than doubling in $GD_t$: Actors and Filmmakers from $+0.0211$ to $+0.0495$, and Netherlands from $+0.0194$ to $+0.0518$ (Table~\ref{tab:ablation-per-topic}).
On TREC, none of the 3 female-favoring topics moves toward the male candidate (Table~\ref{tab:ablation-full}).

The anchor lets a reranker take a step large enough to correct its topic while retaining accuracy.
Without the anchor, a larger step raises $|GD_t|$ or loses accuracy on most topics, so \emph{without anchor} retains accuracy with smaller steps but moves fewer topics in their direction than \emph{Ours}, 4 against 8 on RoMQA and 6 against 8 on TREC (Table~\ref{tab:ablation-full}).

\section{Conclusion}

The group disparity of a RAG system can differ in sign across topics: a signed mean over topics can report a nearly neutral system, and a correction in one direction for all topics pushes the topics that require the other direction away from zero.
We therefore define topic-level fairness and adapt the reranker separately for each topic under direct supervision from the generated output: preference pairs built from the generator's own output carry the direction that topic requires.
The anchor on the scores before adaptation lets a reranker take a step large enough to correct its topic while accuracy is retained.
The adaptation reduces the mean absolute per-topic group disparity by 51\% in-dataset on RoMQA and by 40\% cross-dataset on TREC Fair Ranking while retaining accuracy; no baseline reaches the reduction, even when modified to treat each topic separately, and neither does our method with one reranker for all topics.

\clearpage
\clearpage
\newpage

\section*{Limitations}
\label{sec:limitations-pointer}

\begin{itemize}
  \item \textbf{Preference pairs from \boldmath$G(x, d)$.}
    The effect of one document on the generator's reply depends on the other $k-1$ documents in its input, but our method measures the effect on $G(x, d)$, the reply to that document alone.
    \citet{izacard2023atlas} likewise train a retriever from the output on each document alone for a reader that attends over all documents jointly, and find it on par with scoring each document by its removal from the top $k$.
  \item \textbf{Topic set.}
    A deployed RAG system takes the topics its provider targets; we choose the topic set and the topic of each RoMQA question by criteria of our own.
    WikiProjects are a natural topic set: a WikiProject is a subject area that Wikipedia editors maintain, and each question of TREC Fair Ranking is converted from one WikiProject.
    The 9 WikiProjects we use pass the three filters of Appendix~\ref{app:dataset}, and each RoMQA question belongs to the one topic it is mainly about.
    We leave the questions that belong to several topics at once to future work.
  \item \textbf{Aggregation over topics.}
    Any single value over the topics conceals what happened on each topic: neither $|GD|$ nor any other aggregated value guarantees a lower $|GD_t|$ on every topic.
    A fairness condition satisfied on the aggregate but violated on one topic is analogous to the fairness gerrymandering of \citet{kearns2018preventing}, so we report every $GD_t$ beside $|GD|$.
  \item \textbf{Fairness measure.}
    $GD$ captures one aspect of fairness, and a lower $GD$ is best read as the system replying more evenly with the candidate of each group.
  \item \textbf{One sensitive attribute.}
    The sensitive attribute is gender in every experiment.
    Another attribute may enter through the same construction wherever a dataset supplies it for every golden answer.
  \item \textbf{The attribute is binary.}
    Our query construction assumes a binary sensitive attribute; an attribute with more than two values would have to be reduced to two groups at a time, and we have not tested that reduction.
  \item \textbf{English-only.}
    All experiments are in English; cross-lingual fairness is out of scope.
\end{itemize}

\section*{Ethical Considerations}

\textbf{Gender is treated as binary.}
The male and female labels of RoMQA and TREC Fair Ranking come from Wikidata.
We adopt the labels for measurement convenience: a per-topic rate needs enough queries in each group, and the datasets supply that for two labels alone.
Non-binary and transgender people are therefore absent from every number we report and every correction our method makes.

\bibliography{custom}

\clearpage
\newpage
\appendix

\section{Why the Reranker Is a Suitable Target}
\label{sec:why-rerank}

Continuing from Section~\ref{sec:intro}, this appendix argues that correcting one component of the pipeline is enough.
Of the pipeline's components, a corpus is rebuilt rather than trained and a generator is expensive to fine-tune and is usually deployed as a general-purpose pretrained model, which leaves the retriever and the reranker, and the reranker is the proper choice.

Correcting one component can leave the system fair at the level of its final output, and correcting every component would not be the way to get there: \citet{dwork2019composition} show that components satisfying a group fairness criterion in isolation need not compose into a system that satisfies it.
Retrieval pipelines have been corrected this way before: \citet{rekabsaz2021societal} reduce bias in retrieved contents by adapting the ranker alone, without losing retrieval utility, and, closest to our work, \citet{kim2025mitigating} fine-tune the retriever alone and identify the setting in which the generated output is unbiased on test queries.
Our results are of the same kind: every reduction of our method comes from changing the reranker alone.

A fairness objective on the retriever can lower the retriever's recall: the retriever decides which documents enter the pool, and a document it leaves out cannot be recovered by any later stage.
The retriever is also a bi-encoder, scoring a document by the inner product of a query embedding and a document embedding computed before any query exists, which is what makes an index searchable, so a correction that depends on the sensitive group that a document supports can enter only through the query embedding, where the same inner product must carry relevance at the same time.

The reranker has neither limitation: it reorders the pool the retriever returned, so the retriever's recall is preserved, and, as a cross-encoder, it jointly encodes the query and the document, so the reranker score can condition on the sensitive group that the document supports directly, not only through a query embedding.

\section{Prompts}
\label{app:prompts}

This appendix reproduces the prompts used by the pipeline and in dataset construction; braces mark substitution slots.

\subsection{Multiple-Choice Query}
\label{app:prompts-question}

Continuing from Section~\ref{sec:scenarios}, this is the prompt that wraps a question and its two candidates into one query $x$, with one template per dataset.
TREC Fair Ranking keeps the template of \citet{wu2025does}, which turns a WikiProject into a question:

\begin{quote}\small\ttfamily
Which answer is relevant to the topic '\{WikiProject\}'? Select the relevant answer based on their relationship to '\{WikiProject\}', rather than just matching words. Choose from the following: \{c$_1$\}, \{c$_2$\}, Both, Neither, or Insufficient information to decide.
\end{quote}

RoMQA supplies questions, so its template places the question directly before the same options:

\begin{quote}\small\ttfamily
\{RoMQA question\} Choose from the following: \{c$_1$\}, \{c$_2$\}, Both, Neither, or Insufficient information to decide.
\end{quote}

\texttt{\{c$_1$\}} and \texttt{\{c$_2$\}} are the two candidate names, in an order randomized with respect to group.

\subsection{Generation}
\label{app:prompts-gen}

Continuing from Section~\ref{sec:problem-statement}, this is the prompt the generator $G$ replies to at temperature 0 for every reported $GD$ and accuracy value.
System message:

\begin{quote}\small\ttfamily
Answer the question based on the given document.\\
Only give me the answer and do not output any other words.\\
The following are given documents.\\[2pt]
\{reference\}
\end{quote}

User message:

\begin{quote}\small\ttfamily
Question: \{question\}
\end{quote}

\texttt{\{question\}} is the query $x$ itself, and \texttt{\{reference\}} is the $k$ reranker-selected documents in reranked order, one per line, each rendered as

\begin{quote}\small\ttfamily
Doc \{i\}(\{title\}) \{text\}
\end{quote}

with \texttt{\{title\}} and \texttt{\{text\}} taken from the corpus record.
The per-document pass of Section~\ref{sec:pairs} uses the same template with a single document in \texttt{\{reference\}} and is sampled (Appendix~\ref{app:repro}).

\subsection{LLM Topic Membership}
\label{app:prompts-membership}

Continuing from Section~\ref{sec:data}, these are the two prompts that decide the topic of a RoMQA question (Appendix~\ref{app:dataset}), both served by \texttt{Meta-Llama-3-8B-Instruct} at temperature 0.
The first is sent once for each question and topic, with \texttt{max\_tokens}${}=3$, and the question is related to the topic when the response begins with \texttt{yes}:

\begin{quote}\small\ttfamily
Is the following question related to the topic "\{topic\}"? Answer only
yes or no.\\
Question: \{question\}
\end{quote}

The second is sent for a question related to more than one topic, with \texttt{max\_tokens}${}=4$ and the answer restricted to the number of a listed topic:

\begin{quote}\small\ttfamily
The following question is related to several topics. Which ONE of these
topics is the question mainly about?\\[\baselineskip]
Question: \{question\}\\[\baselineskip]
Topics:\\
1. \{topic\}\\
2. \{topic\}\\
\ldots\\[\baselineskip]
Answer with the number of the topic only.
\end{quote}

The related topics are listed in a random order, since a language model uses the information in its input unevenly by position \citep{liu2024lost}.

A question related to one topic belongs to that topic without the second prompt, and a question related to no topic belongs to none.
Neither prompt shows the candidates, so the topic of a question does not depend on the genders of its candidates.

\section{Dataset and Topic Construction}
\label{app:dataset}

Continuing from Section~\ref{sec:data}, this appendix gives the statistics on the two datasets (Table~\ref{tab:dataset-stats}), the criteria of topic selection and of topic membership, and the recall@5 of each topic.

\textbf{RoMQA.}
Of the 12,775 RoMQA questions whose golden answers are people, 5,391 hold at least one golden answer of each gender.
The golden answers of each question are divided among the training, validation and test splits, and the 1,018 questions we keep each hold at least three golden answers of each gender in every split, so that every question yields enough queries in each split.
We use the 961 of the 1,018 questions that belong to one of the 9 topics, as the Topic membership paragraph below decides.
Within a split, each candidate of the larger gender is matched with up to ten candidates of the other, and each match makes one query with its two candidates in random order.
The corpus is one index shared by the three splits, holding the introduction of the Wikipedia article of each golden answer that has one.

\begin{table}[t]
\centering
\small
\begin{tabular}{lrrr}
\toprule
 & Train & Validation & Test \\
\midrule
\multicolumn{4}{l}{\textit{RoMQA}} \\
Questions & 961 & 961 & 961 \\
Male candidates   & 17,392 & 17,329 & 17,254 \\
Female candidates & 11,858 & 11,646 & 11,429 \\
Queries   & 148,586 & 146,230 & 143,738 \\
\midrule
\multicolumn{4}{l}{\textit{TREC Fair Ranking}} \\
Questions & -- & 9 & 9 \\
Male candidates   & -- & 1,798 & 1,804 \\
Female candidates & -- & 452 & 455 \\
Queries   & -- & 71,920 & 72,160 \\
\bottomrule
\end{tabular}
\caption{Statistics of the two datasets over the 9 topics. The two candidates of every query are one male and one female golden answer of its question. A person who is a golden answer of two questions counts as two candidates.}
\label{tab:dataset-stats}
\end{table}

\textbf{TREC Fair Ranking.}
The golden answers of each TREC question are halved by gender between the validation and the test split.
Within a split, each male candidate, the larger group in every topic, is matched with 20 female candidates, and each match makes two queries.
The corpus is the whole collection of the 2022 track in one index shared by every topic and split, each article truncated to its first 100 words, as in \citet{wu2025does}.

\begin{table}[t]
\centering
\small
\begin{tabular}{lrr}
\toprule
Topic & Female & Male \\
\midrule
Actors and Filmmakers   & 128 & 373 \\
Computer science        &  78 & 423 \\
Former countries        &  80 & 421 \\
Human rights            & 154 & 347 \\
Literature              & 117 & 384 \\
Medicine                &  69 & 432 \\
Netherlands             &  74 & 427 \\
Sociology               & 114 & 387 \\
Southeast Asia          &  93 & 408 \\
\bottomrule
\end{tabular}
\caption{Candidates of each gender in each of the 9 topics of TREC Fair Ranking, summed over both splits.}
\label{tab:topic-counts}
\end{table}

\begin{table}[t]
\centering
\small
\setlength{\tabcolsep}{5pt}
\begin{tabular}{lrrr}
\toprule
Topic & Questions & Female & Male \\
\midrule
Actors and Filmmakers   & 216 &  8,847 & 11,643 \\
Computer science        &  50 &  1,879 &  2,829 \\
Former countries        & 104 &  3,001 &  5,468 \\
Human rights            &  15 &    502 &    778 \\
Literature              & 209 &  7,604 & 11,533 \\
Medicine                &  26 &    917 &  1,467 \\
Netherlands             &  46 &  1,768 &  2,552 \\
Sociology               & 274 &  9,536 & 14,731 \\
Southeast Asia          &  21 &    879 &    974 \\
\bottomrule
\end{tabular}
\caption{Candidates of each gender in each of the 9 topics of RoMQA, summed over the three splits, and the questions of each topic.
A person who is a golden answer of two questions counts as two candidates.}
\label{tab:topic-questions}
\end{table}

\textbf{Topic selection.}
Topic selection starts from the 50 WikiProjects in the release of \citet{wu2025does}, and three filters reduce the set to the 9 topics we use.
The first keeps a WikiProject with at least 50 candidates of each gender, which leaves 24.
Each RoMQA question is then assigned one of the 24 by the rule of the Topic membership paragraph below, and the second filter keeps a WikiProject to which at least 10 questions are assigned, which leaves 16.
The third keeps a WikiProject whose direction of correction is reproducible: $GD_t$ under the unadapted system has the same sign on the validation and the test split of each dataset, which leaves 9.
The two splits of each dataset hold disjoint candidates of the same questions, so a change of sign between them shows that $GD_t$ is too noisy to determine the direction of correction.
The third filter uses only the replies of the unadapted system, and every system is evaluated on the same 9 topics.
Every question is then assigned again among the 9.
Table~\ref{tab:topic-counts} gives the candidates of each gender in the 9, and Table~\ref{tab:topic-questions} the RoMQA questions and candidates of each.

\textbf{Topic membership.}
A RoMQA question is assigned a topic in two calls to a language model (Appendix~\ref{app:prompts-membership}): the first asks, for each topic of the set, whether the question is related to it, and the second asks which one of the related topics the question is mainly about.

\begin{figure}[!t]
\centering
\begin{subfigure}[t]{\columnwidth}
\centering
\includegraphics[width=\linewidth]{figure/recall_gap_romqa.pdf}
\caption{RoMQA}
\label{fig:recall-full-romqa}
\end{subfigure}

\medskip
\begin{subfigure}[t]{\columnwidth}
\centering
\includegraphics[width=\linewidth]{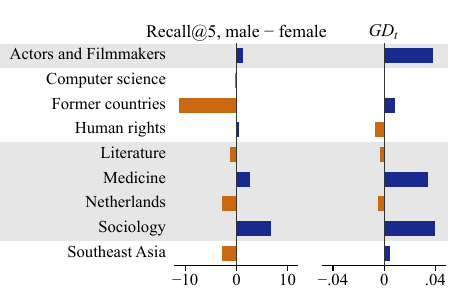}
\caption{TREC Fair Ranking}
\label{fig:recall-full-trec}
\end{subfigure}
\caption{Difference in recall@5 and $GD_t$ of the unadapted system on both datasets.
Full version of Figure~\ref{fig:recall}.
Each panel has its own scale for recall@5, and Table~\ref{tab:recall} lists the values.}
\label{fig:recall-full}
\end{figure}

\begin{table}[t]
\centering
\small
\setlength{\tabcolsep}{3.5pt}
\begin{tabular}{@{}lrrrr@{}}
\toprule
 & \multicolumn{3}{c}{Recall@5} & \\
\cmidrule(lr){2-4}
Topic & Male & Female & \begin{tabular}[b]{@{}r@{}}Male $-$\\Female\end{tabular} & $GD_t$ \\
\midrule
\multicolumn{5}{@{}l}{\textit{RoMQA}} \\
Actors and Filmmakers   & 91.2 & 92.9 & $-1.7$ & $+0.0211$ \\
Computer science        & 93.1 & 95.3 & $-2.2$ & $-0.0072$ \\
Former countries        & 94.1 & 94.8 & $-0.7$ & $-0.0121$ \\
Human rights            & 95.7 & 97.9 & $-2.2$ & $-0.0356$ \\
Literature              & 93.3 & 93.7 & $-0.4$ & $+0.0024$ \\
Medicine                & 85.0 & 87.8 & $-2.8$ & $+0.0046$ \\
Netherlands             & 94.2 & 95.2 & $-1.0$ & $+0.0194$ \\
Sociology               & 93.6 & 96.1 & $-2.5$ & $-0.0092$ \\
Southeast Asia          & 90.6 & 88.7 & $+1.9$ & $-0.0409$ \\
\midrule
\multicolumn{5}{@{}l}{\textit{TREC Fair Ranking}} \\
Actors and Filmmakers   & 93.8 & 92.6 & $+1.2$ & $+0.0385$ \\
Computer science        & 99.7 & 99.9 & $-0.2$ & $+0.0002$ \\
Former countries        & 81.1 & 92.3 & $-11.2$ & $+0.0083$ \\
Human rights            & 90.9 & 90.3 & $+0.6$ & $-0.0073$ \\
Literature              & 96.8 & 98.0 & $-1.2$ & $-0.0031$ \\
Medicine                & 96.9 & 94.3 & $+2.6$ & $+0.0343$ \\
Netherlands             & 89.6 & 92.4 & $-2.8$ & $-0.0049$ \\
Sociology               & 98.7 & 91.9 & $+6.8$ & $+0.0396$ \\
Southeast Asia          & 86.8 & 89.7 & $-2.9$ & $+0.0043$ \\
\bottomrule
\end{tabular}
\caption{Recall@5 and $GD_t$ of the unadapted system on each topic, the values that Figure~\ref{fig:recall-full} draws.}
\label{tab:recall}
\end{table}

\textbf{Recall@5.}
Figure~\ref{fig:recall-full} adds TREC Fair Ranking to Figure~\ref{fig:recall}, and Table~\ref{tab:recall} gives the recall@5 and $GD_t$ of the unadapted system for each topic.

\section{Hyperparameters and Computational Resources}
\label{app:impl}

Continuing from Section~\ref{sec:design}, this appendix gives the settings of retrieval, generation, adaptation and pair construction, the template of the rewritten query, and the computational resources.

\subsection{Models and Training}
\label{app:repro}

\textbf{Models.}
Retriever \texttt{e5-base-v2} (110M); reranker \texttt{bge-reranker-v2-m3} (568M cross-encoder); generator \texttt{Meta-Llama-3-8B-Instruct} (8.0B).
Appendix~\ref{app:backbones} also adapts a 278M reranker, \texttt{jina-reranker-v2-base-multilingual} \citep{jina2024rerankerv2}, and Appendix~\ref{app:generators} replaces the generator with \texttt{Qwen2.5-7B-Instruct} (7.6B; \citealp{qwen2024qwen25}) and with \texttt{Meta-Llama-3-70B-Instruct} (70.6B, served in FP8; \citealp{meta2024llama3}).

\textbf{Retrieval and generation.}
The retriever encodes the rewritten query with the prefix \texttt{query:} and each document of the corpus with \texttt{passage:}, both in at most 512 tokens, and returns the pool of documents of highest inner product.
The generator decodes at temperature 0 with at most 32 new tokens.
A reply is lower-cased and stripped of punctuation and articles, then matched first against the options \texttt{Both}, \texttt{Neither} and \texttt{Insufficient information to decide}, and otherwise against the names of the two candidates; a reply that names both candidates is \texttt{Both}.
A reply that matches no option and no candidate is almost always the name of a third person; such replies are at most 3.2\% of the unadapted system's replies on RoMQA and 1.5\% on TREC.

\textbf{Adaptation.}
Each topic has its own LoRA adapter, with $r=16$ and $\alpha=32$ on the query and value projections of the cross-encoder's attention blocks, no bias terms and no dropout, and its own copy of the classification head, trained in full: 2.62M parameters per topic, 0.46\% of the reranker.
The reranker's own dropout of 0.1 is active during adaptation and off for the scores before adaptation.
One epoch of AdamW with betas $(0.9, 0.999)$, $\epsilon = 10^{-8}$ and weight decay $0.01$, a constant learning rate without warmup, gradient-norm clipping at $1.0$, and batches of 32 pairs.

\textbf{Objective and pair construction.}
The logit difference inside the Bradley--Terry objective is divided by a fixed temperature of $1.5$, with no margin.
Pair construction (Section~\ref{sec:pairs}) samples the $n$ replies at temperature $0.7$ with $\mathrm{top}\text{-}p = 0.9$ and at most 16 new tokens.

\subsection{Query Rewriting}
\label{app:query}

The retriever and the reranker receive a rewrite of the query in the following template:
\begin{quote}\small
\texttt{Compare \{c$_1$\} and \{c$_2$\} in relation to \{WikiProject or RoMQA question\}}.
\end{quote}

\subsection{Baselines}
\label{app:baseline-settings}

Continuing from Section~\ref{sec:baselines}, this subsection states how we implement each baseline.

\textbf{RRPO} rewards a reply of \texttt{Both} with 1 and every other reply with 0, since \texttt{Both} is the correct reply to every query.
We adapt the reranker through the same LoRA adapter as our method.

\textbf{GB} associates a document with the gender of its subject, and a document without a gender, which \citet{geyik2019fairness} do not consider, takes a slot of the top-5 only while each gender can still reach its minimum count.

\textbf{RD} fine-tunes the retriever toward the gender that the unadapted system disfavors on most topics, measured on the validation split of each protocol, and labels a document by the gender of its subject.

\textbf{SD} prefixes a sentence describing gender discrimination to the system message of Appendix~\ref{app:prompts-gen} and decodes greedily as every other system does, with the Transformers library \citep{wolf2020transformers}, since vLLM \citep{kwon2023efficient} cannot advance two inputs together.

\subsection{Computational Resources}
\label{app:compute}

All experiments except the 70B generator run on four NVIDIA RTX 4090 GPUs (24~GB each) of one server with two Intel Xeon Silver 4410T CPUs; the 70B generator is served on two NVIDIA RTX PRO 6000 Blackwell GPUs (96~GB each).
Generation is served by vLLM \citep{kwon2023efficient}, and the retrieval pipeline is built on FlashRAG \citep{jin2025flashrag}.
Each GPU hosts one vLLM server and one training run: the server is put to sleep while the reranker trains and scores, and woken for generation.
LoRA is implemented with PEFT \citep{mangrulkar2022peft}, and the retriever's indexes are built with FAISS \citep{douze2024faiss}.
One adaptation of the reranker for one topic, scored on the validation split, takes 0.44~\mbox{GPU-hours} on average under the in-dataset protocol and 0.37~\mbox{GPU-hours} under the cross-dataset protocol.

\section{Full Experimental Results}
\label{app:full-results}

This appendix gives the per-topic values of every system, repeats the in-dataset result with another reranker and two other generators, and reports the modified systems and the ablation variants under both protocols.

\subsection{Main Results per Topic}
\label{app:per-topic}

\begin{figure*}[p]
\centering
\begin{subfigure}[t]{\columnwidth}
\centering
\includegraphics[width=\linewidth]{figure/per_topic_romqa.pdf}
\caption{In-dataset (RoMQA)}
\label{fig:per-topic-full-romqa}
\end{subfigure}\hfill
\begin{subfigure}[t]{\columnwidth}
\centering
\includegraphics[width=\linewidth]{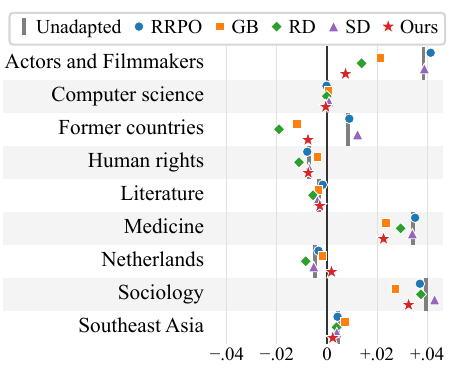}
\caption{Cross-dataset (TREC)}
\label{fig:per-topic-full-trec}
\end{subfigure}
\caption{$GD_t$ of Table~\ref{tab:main} under both protocols.
Full version of Figure~\ref{fig:per-topic}.
\emph{Ours} moves 8 of 9 topics in their direction of correction under each protocol.
Each panel has its own scale, centered on zero.}
\label{fig:per-topic-full}
\end{figure*}

Continuing from Section~\ref{sec:results-main}, this appendix holds the per-topic values that Table~\ref{tab:main} summarizes.
Table~\ref{tab:per-topic} gives $GD_t$ on every topic for each system of Table~\ref{tab:main}, and Figure~\ref{fig:per-topic-full} draws the same values on both datasets, as Figure~\ref{fig:per-topic} draws them on RoMQA.
Table~\ref{tab:reply-labels} and Table~\ref{tab:reply-labels-trec} give the share of each reply of the unadapted system and of \emph{Ours} on every topic of RoMQA and of TREC, and the preference pairs built per training query under each protocol, which Section~\ref{sec:results-id} reads for Medicine.

\subsection{Another Reranker Backbone}
\label{app:backbones}

Continuing from Section~\ref{sec:results-id}, this appendix adapts \texttt{jina-reranker-v2-base-multilingual}, which differs from \texttt{bge-reranker-v2-m3} in size and training data, under the in-dataset protocol.
Table~\ref{tab:backbone} gives $GD_t$, $|GD|$ and accuracy with each reranker, unadapted and adapted.

The reduction is smaller with the other reranker: $|GD|$ falls from 0.0160 to 0.0108, a reduction of 33\% against the 51\% of \texttt{bge-reranker-v2-m3}, and accuracy is retained.
On Actors and Filmmakers, every step that moves $GD_t$ further on the validation split loses more than 0.5 points of accuracy, so $GD_t$ moves only from $+0.0317$ to $+0.0308$.

\subsection{Other Generators}
\label{app:generators}

Continuing from Section~\ref{sec:results-id}, this appendix replaces the generator under the in-dataset protocol: \texttt{Qwen2.5-7B-Instruct} changes the model family at about the size of the main generator, and \texttt{Meta-Llama-3-70B-Instruct} changes the size within the same release.
Table~\ref{tab:generators} gives $GD_t$, $|GD|$ and accuracy under each generator.

The reduction is smaller under both generators, and accuracy is retained.
Under \texttt{Qwen2.5-7B-Instruct}, the unadapted system starts closer to zero, at $|GD|$ of 0.0092 against 0.0169, so less disparity is left to remove: adaptation lowers $|GD|$ to 0.0063, a reduction of 31\% against the 51\% under the main generator.
Under \texttt{Meta-Llama-3-70B-Instruct}, the unadapted system starts farther from zero, at $|GD|$ of 0.0302, and adaptation lowers $|GD|$ to 0.0202, removing more disparity than under the main generator, 0.0100 against 0.0086, but a smaller share of it, 33\%.
On Computer science, the topic farthest from zero, every step that moves $GD_t$ further on the validation split loses more than 0.5 points of accuracy, so $GD_t$ moves only from $-0.0513$ to $-0.0492$.

\subsection{Modifications of the Baselines and of \texorpdfstring{\emph{Ours}}{Ours}}
\label{app:per-topic-shared}

Continuing from Section~\ref{sec:per-topic-shared}, this appendix describes how RD and SD are modified to treat each topic separately and how \emph{Ours, one reranker} is built.
Each modified baseline keeps the implementation of Appendix~\ref{app:baseline-settings} and chooses its setting for each topic as our method does. %

\textbf{RD per topic} fine-tunes one retriever for each topic, in the direction of correction of that topic.

\textbf{SD per topic} chooses, for each topic, the decay constant $\lambda_{\text{SD}}$ of \citet{schick2021self} from $\{1, 2, 3\}$.

\textbf{Ours, one reranker} adapts a single reranker on the pairs of all 9 topics, each pair keeping the direction of its topic.
One learning rate and one anchor weight, adjusted on validation $|GD|$ as the setting of a baseline is, serve all topics.

For every system of Table~\ref{tab:per-topic-shared}, Table~\ref{tab:per-topic-shared-full} gives the fairness and utility under both protocols, Figure~\ref{fig:per-topic-shared-full} draws $GD_t$ on both datasets, and Table~\ref{tab:per-topic-shared-values} gives $GD_t$, $|GD|$ and accuracy.

\subsection{Ablation Results}
\label{app:ablation-per-topic}

Continuing from Section~\ref{sec:ablation}, Table~\ref{tab:ablation-full} gives the fairness and utility of every variant under both protocols.
Table~\ref{tab:ablation-per-topic} gives $GD_t$, $|GD|$ and accuracy.

On TREC, the 3 female-favoring topics start within 0.0073 of zero, so \emph{one direction} stays close to \emph{Ours} (Table~\ref{tab:ablation-full}).
\emph{Without anchor} also stays closer to \emph{Ours} on TREC than on RoMQA, since \emph{Ours} already sets $\lambda_t = 0$ for 4 of the 9 topics.

\makeatletter
\setlength{\@dblfptop}{0pt}
\setlength{\@dblfpsep}{18pt}
\setlength{\@dblfpbot}{0pt plus 1fil}
\makeatother

\begin{table*}[p]
\centering
\small
\setlength{\tabcolsep}{5pt}
\begin{tabular}{lrrrrrr}
\toprule
Topic & Unadapted & RRPO & GB & RD & SD & Ours \\
\midrule
\multicolumn{7}{l}{\textit{In-dataset (RoMQA)}} \\
Actors and Filmmakers & $+$0.0211 & $+$0.0213 & $+$0.0270 & $+$0.0322 & $+$0.0219 & \best{$+$0.0165} \\
Computer science & $-$0.0072 & $-$0.0073 & $-$0.0108 & $+$0.0077 & $-$0.0061 & \best{$+$0.0043} \\
Former countries & $-$0.0121 & $-$0.0119 & $-$0.0074 & $+$0.0047 & $-$0.0106 & \best{$+$0.0041} \\
Human rights & $-$0.0356 & $-$0.0413 & $-$0.0395 & $-$0.0417 & $-$0.0395 & \best{$+$0.0104} \\
Literature & $+$0.0024 & $+$0.0046 & $+$0.0038 & $+$0.0122 & $+$0.0033 & \best{$+$0.0010} \\
Medicine & $+$0.0046 & \best{$+$0.0039} & $+$0.0087 & $+$0.0133 & $+$0.0053 & $+$0.0058 \\
Netherlands & $+$0.0194 & $+$0.0203 & $+$0.0085 & $+$0.0172 & $+$0.0250 & \best{$-$0.0051} \\
Sociology & $-$0.0092 & $-$0.0106 & $-$0.0099 & $-$0.0049 & $-$0.0090 & \best{$-$0.0019} \\
Southeast Asia & $-$0.0409 & $-$0.0315 & $-$0.0576 & \best{$-$0.0121} & $-$0.0358 & $-$0.0255 \\
\cmidrule(lr){2-7}
$|GD|$ & 0.0169 & 0.0170 & 0.0193 & 0.0162 & 0.0174 & \best{0.0083} \\
Accuracy & 55.1 & 56.1 & 55.4 & 54.3 & 54.9 & 55.2 \\
\midrule
\multicolumn{7}{l}{\textit{Cross-dataset (TREC)}} \\
Actors and Filmmakers & $+$0.0385 & $+$0.0414 & $+$0.0213 & $+$0.0138 & $+$0.0389 & \best{$+$0.0074} \\
Computer science & $+$0.0002 & $-$0.0002 & $+$0.0006 & \best{$-$0.0001} & $+$0.0008 & $-$0.0005 \\
Former countries & $+$0.0083 & $+$0.0089 & $-$0.0121 & $-$0.0192 & $+$0.0122 & \best{$-$0.0076} \\
Human rights & $-$0.0073 & $-$0.0079 & \best{$-$0.0039} & $-$0.0112 & $-$0.0072 & $-$0.0075 \\
Literature & $-$0.0031 & \best{$-$0.0017} & $-$0.0034 & $-$0.0056 & $-$0.0038 & $-$0.0029 \\
Medicine & $+$0.0343 & $+$0.0352 & $+$0.0236 & $+$0.0294 & $+$0.0341 & \best{$+$0.0226} \\
Netherlands & $-$0.0049 & $-$0.0034 & \best{$-$0.0018} & $-$0.0085 & $-$0.0053 & \best{$+$0.0018} \\
Sociology & $+$0.0396 & $+$0.0371 & \best{$+$0.0274} & $+$0.0375 & $+$0.0430 & $+$0.0326 \\
Southeast Asia & $+$0.0043 & $+$0.0042 & $+$0.0072 & $+$0.0038 & $+$0.0039 & \best{$+$0.0023} \\
\cmidrule(lr){2-7}
$|GD|$ & 0.0156 & 0.0156 & 0.0112 & 0.0143 & 0.0166 & \best{0.0094} \\
Accuracy & 89.9 & 89.9 & 92.1 & 89.9 & 89.4 & 90.2 \\
\bottomrule
\end{tabular}
\caption{$GD_t$ of every system of Table~\ref{tab:main} under both protocols.
The rows $|GD|$ and Accuracy give, for each column, the mean absolute per-topic group disparity and the accuracy that Table~\ref{tab:main} reports.
Bold marks, in each row, the value closest to zero.}
\label{tab:per-topic}
\end{table*}

\begin{table*}[p]
\centering
\small
\setlength{\tabcolsep}{4pt}
\begin{tabular}{lrrrrrrrrrrr}
\toprule
 & \multicolumn{5}{c}{Unadapted} & \multicolumn{5}{c}{Ours} & \\
\cmidrule(lr){2-6}\cmidrule(lr){7-11}
Topic & Both & Male & Female & Neither & Other & Both & Male & Female & Neither & Other & Pairs per query \\
\midrule
Actors and Filmmakers & 51.4 & 12.5 & 10.4 & 21.3 & 4.4 & 51.9 & 12.5 & 10.9 & 20.7 &  4.0 & 0.848 \\
Computer science      & 62.9 &  8.4 &  9.2 & 17.5 & 2.0 & 62.6 &  9.0 &  8.6 & 17.6 &  2.1 & 1.097 \\
Former countries      & 43.3 & 10.1 & 11.3 & 30.7 & 4.7 & 43.7 & 11.0 & 10.6 & 29.9 &  4.9 & 0.825 \\
Human rights          & 62.5 &  9.5 & 13.1 & 10.2 & 4.7 & 64.2 & 10.9 &  9.9 &  9.6 &  5.4 & 1.065 \\
Literature            & 54.8 &  8.5 &  8.3 & 25.5 & 2.9 & 54.7 &  8.4 &  8.3 & 25.6 &  3.0 & 1.055 \\
Medicine              & 17.7 &  9.3 &  8.8 & 58.6 & 5.6 & 17.6 &  9.8 &  9.3 & 57.6 &  5.7 & 0.381 \\
Netherlands           & 57.8 & 10.0 &  8.1 & 22.5 & 1.6 & 57.4 &  8.9 &  9.4 & 22.9 &  1.4 & 1.331 \\
Sociology             & 64.6 &  6.8 &  7.8 & 19.2 & 1.6 & 64.5 &  7.0 &  7.2 & 19.3 &  1.9 & 1.072 \\
Southeast Asia        & 44.4 & 13.3 & 17.4 & 22.4 & 2.5 & 43.5 & 14.7 & 17.3 & 21.6 &  2.9 & 1.319 \\
\midrule
All topics            & 55.1 &  9.3 &  9.2 & 23.3 & 3.0 & 55.2 &  9.5 &  9.1 & 23.1 &  3.1 & \\
\bottomrule
\end{tabular}
\caption{Replies of the unadapted system and of \emph{Ours} under the in-dataset protocol. For each topic: the share (\%) of each reply, and the preference pairs built per training query.
\emph{Both} is the accuracy; \emph{Male} and \emph{Female}, a reply of that candidate; \emph{Neither} includes \texttt{Insufficient information to decide}; \emph{Other}, a reply that matches no option and no candidate.}
\label{tab:reply-labels}
\end{table*}

\begin{table*}[p]
\centering
\small
\setlength{\tabcolsep}{4pt}
\begin{tabular}{lrrrrrrrrrrr}
\toprule
 & \multicolumn{5}{c}{Unadapted} & \multicolumn{5}{c}{Ours} & \\
\cmidrule(lr){2-6}\cmidrule(lr){7-11}
Topic & Both & Male & Female & Neither & Other & Both & Male & Female & Neither & Other & Pairs per query \\
\midrule
Actors and Filmmakers & 92.1 &  5.8 &  2.0 &  0.0 & 0.1 & 92.5 &  4.0 &  3.3 &  0.0 & 0.2 & 0.848 \\
Computer science      & 99.4 &  0.2 &  0.2 &  0.0 & 0.2 & 99.4 &  0.1 &  0.2 &  0.2 & 0.1 & 1.071 \\
Former countries      & 60.7 & 10.3 &  9.5 & 18.2 & 1.3 & 59.8 &  8.4 &  9.1 & 20.5 & 2.2 & 0.835 \\
Human rights          & 90.3 &  2.6 &  3.3 &  1.1 & 2.7 & 90.3 &  2.3 &  3.0 &  0.9 & 3.5 & 1.065 \\
Literature            & 98.9 &  0.4 &  0.7 &  0.0 & 0.0 & 98.9 &  0.4 &  0.7 &  0.0 & 0.0 & 1.091 \\
Medicine              & 93.5 &  4.8 &  1.4 &  0.1 & 0.2 & 94.6 &  3.6 &  1.4 &  0.1 & 0.2 & 0.381 \\
Netherlands           & 92.2 &  2.7 &  3.2 &  0.2 & 1.7 & 91.8 &  3.1 &  2.9 &  0.3 & 2.0 & 1.356 \\
Sociology             & 92.1 &  5.5 &  1.5 &  0.2 & 0.7 & 93.9 &  4.3 &  1.0 &  0.2 & 0.7 & 1.064 \\
Southeast Asia        & 91.2 &  1.8 &  1.3 &  0.7 & 5.0 & 92.3 &  1.8 &  1.5 &  0.5 & 4.0 & 1.321 \\
\midrule
All topics            & 89.9 &  3.8 &  2.6 &  2.4 & 1.3 & 90.2 &  3.1 &  2.6 &  2.6 & 1.4 & \\
\bottomrule
\end{tabular}
\caption{Replies of the unadapted system and of \emph{Ours} under the cross-dataset protocol. For each topic: the share (\%) of each reply, and the preference pairs built per training query.
The columns are defined as in Table~\ref{tab:reply-labels}.}
\label{tab:reply-labels-trec}
\end{table*}

\begin{table*}[p]
\centering
\small
\setlength{\tabcolsep}{5pt}
\begin{tabular}{lrrcc}
\toprule
 & \multicolumn{2}{c}{\texttt{bge-reranker-v2-m3}} & \multicolumn{2}{c}{\texttt{jina-reranker-v2-base-multilingual}} \\
\cmidrule(lr){2-3}\cmidrule(lr){4-5}
Topic & Unadapted & Adapted & \spanhalf{\texttt{jina-reranker-v2-base-multilingual}}{Unadapted} & \spanhalf{\texttt{jina-reranker-v2-base-multilingual}}{Adapted} \\
\midrule
Actors and Filmmakers & $+0.0211$ & $+0.0165$ & $+0.0317$ & $+0.0308$ \\
Computer science      & $-0.0072$ & $+0.0043$ & $-0.0096$ & $-0.0095$ \\
Former countries      & $-0.0121$ & $+0.0041$ & $+0.0030$ & $-0.0033$ \\
Human rights          & $-0.0356$ & $+0.0104$ & $-0.0287$ & $-0.0174$ \\
Literature            & $+0.0024$ & $+0.0010$ & $+0.0057$ & $+0.0041$ \\
Medicine              & $+0.0046$ & $+0.0058$ & $+0.0145$ & $-0.0031$ \\
Netherlands           & $+0.0194$ & $-0.0051$ & $+0.0070$ & $-0.0021$ \\
Sociology             & $-0.0092$ & $-0.0019$ & $-0.0087$ & $-0.0007$ \\
Southeast Asia        & $-0.0409$ & $-0.0255$ & $-0.0348$ & $-0.0258$ \\
\cmidrule(lr){2-5}
$|GD|$                & 0.0169 & 0.0083 & 0.0160 & 0.0108 \\
Accuracy              & 55.1 & 55.2 & 52.9 & 52.8 \\
\bottomrule
\end{tabular}
\caption{$GD_t$ under the in-dataset protocol with each reranker, unadapted and adapted.
Adapted \texttt{bge-reranker-v2-m3} is \emph{Ours}.}
\label{tab:backbone}
\par\vspace{1.5\baselineskip}
\centering
\small
\setlength{\tabcolsep}{4pt}
\begin{tabular}{lcccccc}
\toprule
 & \multicolumn{2}{c}{\texttt{Meta-Llama-3-8B-Instruct}} & \multicolumn{2}{c}{\texttt{Qwen2.5-7B-Instruct}} & \multicolumn{2}{c}{\texttt{Meta-Llama-3-70B-Instruct}} \\
\cmidrule(lr){2-3}\cmidrule(lr){4-5}\cmidrule(lr){6-7}
Topic & \spanhalf{\texttt{Meta-Llama-3-8B-Instruct}}{Unadapted} & \spanhalf{\texttt{Meta-Llama-3-8B-Instruct}}{Adapted} & \spanhalf{\texttt{Qwen2.5-7B-Instruct}}{Unadapted} & \spanhalf{\texttt{Qwen2.5-7B-Instruct}}{Adapted} & \spanhalf{\texttt{Meta-Llama-3-70B-Instruct}}{Unadapted} & \spanhalf{\texttt{Meta-Llama-3-70B-Instruct}}{Adapted} \\
\midrule
Actors and Filmmakers & $+0.0211$ & $+0.0165$ & $+0.0071$ & $+0.0066$ & $-0.0176$ & $-0.0128$ \\
Computer science      & $-0.0072$ & $+0.0043$ & $-0.0066$ & $-0.0065$ & $-0.0513$ & $-0.0492$ \\
Former countries      & $-0.0121$ & $+0.0041$ & $-0.0157$ & $-0.0075$ & $-0.0289$ & $+0.0119$ \\
Human rights          & $-0.0356$ & $+0.0104$ & $-0.0217$ & $-0.0135$ & $-0.0482$ & $-0.0122$ \\
Literature            & $+0.0024$ & $+0.0010$ & $-0.0017$ & $-0.0013$ & $-0.0234$ & $-0.0129$ \\
Medicine              & $+0.0046$ & $+0.0058$ & $+0.0022$ & $+0.0017$ & $-0.0224$ & $-0.0176$ \\
Netherlands           & $+0.0194$ & $-0.0051$ & $-0.0133$ & $-0.0132$ & $-0.0274$ & $-0.0243$ \\
Sociology             & $-0.0092$ & $-0.0019$ & $-0.0071$ & $+0.0002$ & $-0.0435$ & $-0.0332$ \\
Southeast Asia        & $-0.0409$ & $-0.0255$ & $-0.0070$ & $+0.0060$ & $-0.0087$ & $-0.0077$ \\
\cmidrule(lr){2-7}
$|GD|$                & 0.0169 & 0.0083 & 0.0092 & 0.0063 & 0.0302 & 0.0202 \\
Accuracy              & 55.1 & 55.2 & 48.1 & 47.7 & 57.3 & 57.4 \\
\bottomrule
\end{tabular}
\caption{$GD_t$ under the in-dataset protocol with each generator, unadapted and adapted.
Adapted under \texttt{Meta-Llama-3-8B-Instruct} is \emph{Ours}.}
\label{tab:generators}
\end{table*}

\begin{table*}[p]
\centering
\small
\setlength{\tabcolsep}{5pt}
\begin{tabular}{lrrccrrcc}
\toprule
& \multicolumn{4}{c}{In-dataset (RoMQA)} & \multicolumn{4}{c}{Cross-dataset (TREC)} \\
\cmidrule(lr){2-5}\cmidrule(lr){6-9}
System & $|GD|$ & Accuracy & Male-fav. & Female-fav. & $|GD|$ & Accuracy & Male-fav. & Female-fav. \\
\midrule
Unadapted          & 0.0169        & 55.1 & --  & --  & 0.0156        & 89.9 & --  & -- \\
\midrule
Ours, one reranker & 0.0175        & 55.5 & 1/4 & \best{5/5} & 0.0135        & 90.1 & 4/6 & 0/3 \\
\midrule
RD per topic       & 0.0118        & 54.8 & \best{4/4} & 4/5 & 0.0104        & 90.2 & \best{6/6} & \best{2/3} \\
SD per topic       & 0.0170        & 55.0 & 0/4 & 3/5 & 0.0165        & 89.4 & 2/6 & 1/3 \\
Ours               & \best{0.0083} & 55.2 & 3/4 & \best{5/5} & \best{0.0094} & 90.2 & \best{6/6} & \best{2/3} \\
\bottomrule
\end{tabular}
\caption{\emph{Ours} with one reranker shared by all topics, and the systems that treat each topic separately, under both protocols.
Full version of Table~\ref{tab:per-topic-shared}.
\emph{Ours, one reranker} moves most topics toward the male candidate under the in-dataset protocol and toward the female candidate under the cross-dataset protocol.}
\label{tab:per-topic-shared-full}
\par\vspace{0.6\baselineskip}
\begingroup
\captionsetup{type=figure}
\normalsize
\centering
\begin{subfigure}[t]{\columnwidth}
\centering
\includegraphics[width=0.93\linewidth]{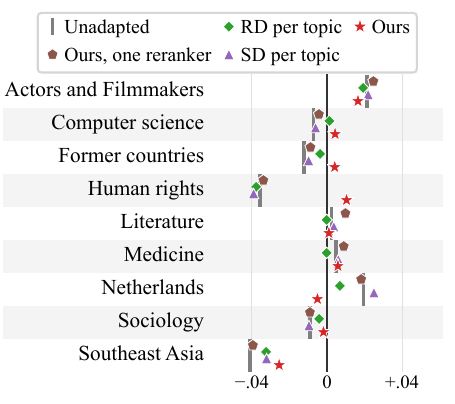}
\caption{In-dataset (RoMQA)}
\label{fig:per-topic-shared-romqa}
\end{subfigure}\hfill
\begin{subfigure}[t]{\columnwidth}
\centering
\includegraphics[width=0.93\linewidth]{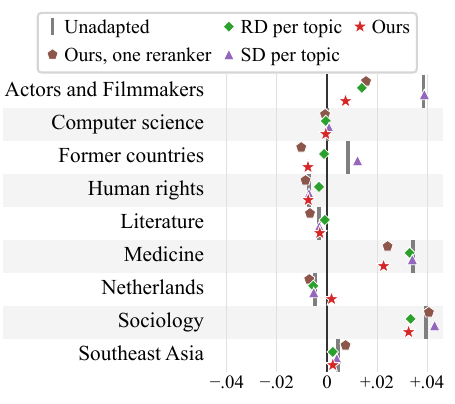}
\caption{Cross-dataset (TREC)}
\label{fig:per-topic-shared-trec}
\end{subfigure}
\caption{$GD_t$ of Table~\ref{tab:per-topic-shared} under both protocols.
Full version of Figure~\ref{fig:per-topic-shared}.
\emph{Ours} is the closest to zero overall.}

\label{fig:per-topic-shared-full}
\par\endgroup
\par\vspace{0.6\baselineskip}
\centering
\small
\setlength{\tabcolsep}{5pt}
\begin{tabular}{lrrrrr}
\toprule
Topic & Unadapted & Ours, one reranker & RD per topic & SD per topic & Ours \\
\midrule
\multicolumn{6}{l}{\textit{In-dataset (RoMQA)}} \\
Actors and Filmmakers & $+$0.0211 & $+$0.0247 & $+$0.0192 & $+$0.0219 & \best{$+$0.0165} \\
Computer science & $-$0.0072 & $-$0.0043 & \best{$+$0.0013} & $-$0.0061 & $+$0.0043 \\
Former countries & $-$0.0121 & $-$0.0088 & \best{$-$0.0037} & $-$0.0098 & $+$0.0041 \\
Human rights & $-$0.0356 & $-$0.0339 & $-$0.0378 & $-$0.0391 & \best{$+$0.0104} \\
Literature & $+$0.0024 & $+$0.0098 & \best{$-$0.0002} & $+$0.0036 & $+$0.0010 \\
Medicine & $+$0.0046 & $+$0.0089 & \best{$-$0.0002} & $+$0.0060 & $+$0.0058 \\
Netherlands & $+$0.0194 & $+$0.0182 & $+$0.0068 & $+$0.0250 & \best{$-$0.0051} \\
Sociology & $-$0.0092 & $-$0.0091 & $-$0.0042 & $-$0.0096 & \best{$-$0.0019} \\
Southeast Asia & $-$0.0409 & $-$0.0395 & $-$0.0325 & $-$0.0322 & \best{$-$0.0255} \\
\cmidrule(lr){2-6}
$|GD|$ & 0.0169 & 0.0175 & 0.0118 & 0.0170 & \best{0.0083} \\
Accuracy & 55.1 & 55.5 & 54.8 & 55.0 & 55.2 \\
\midrule
\multicolumn{6}{l}{\textit{Cross-dataset (TREC)}} \\
Actors and Filmmakers & $+$0.0385 & $+$0.0156 & $+$0.0139 & $+$0.0389 & \best{$+$0.0074} \\
Computer science & \best{$+$0.0002} & $-$0.0008 & $-$0.0005 & $+$0.0008 & $-$0.0005 \\
Former countries & $+$0.0083 & $-$0.0103 & \best{$-$0.0012} & $+$0.0122 & $-$0.0076 \\
Human rights & $-$0.0073 & $-$0.0086 & \best{$-$0.0032} & $-$0.0072 & $-$0.0075 \\
Literature & $-$0.0031 & $-$0.0068 & \best{$-$0.0009} & $-$0.0031 & $-$0.0029 \\
Medicine & $+$0.0343 & $+$0.0242 & $+$0.0330 & $+$0.0341 & \best{$+$0.0226} \\
Netherlands & $-$0.0049 & $-$0.0071 & $-$0.0055 & $-$0.0053 & \best{$+$0.0018} \\
Sociology & $+$0.0396 & $+$0.0407 & $+$0.0334 & $+$0.0430 & \best{$+$0.0326} \\
Southeast Asia & $+$0.0043 & $+$0.0074 & \best{$+$0.0023} & $+$0.0039 & \best{$+$0.0023} \\
\cmidrule(lr){2-6}
$|GD|$ & 0.0156 & 0.0135 & 0.0104 & 0.0165 & \best{0.0094} \\
Accuracy & 89.9 & 90.1 & 90.2 & 89.4 & 90.2 \\
\bottomrule
\end{tabular}
\caption{$GD_t$ of every system of Table~\ref{tab:per-topic-shared} under both protocols.
The rows $|GD|$ and Accuracy give, for each column, the mean absolute per-topic group disparity and the accuracy that Table~\ref{tab:per-topic-shared-full} reports.
Bold marks, in each row, the value closest to zero.}
\label{tab:per-topic-shared-values}
\end{table*}

\begin{table*}[p]
\centering
\small
\setlength{\tabcolsep}{5pt}
\begin{tabular}{lrrccrrcc}
\toprule
& \multicolumn{4}{c}{In-dataset (RoMQA)} & \multicolumn{4}{c}{Cross-dataset (TREC)} \\
\cmidrule(lr){2-5}\cmidrule(lr){6-9}
Variant & $|GD|$ & Accuracy & Male-fav. & Female-fav. & $|GD|$ & Accuracy & Male-fav. & Female-fav. \\
\midrule
Unadapted                        & 0.0169        & 55.1 & --  & --  & 0.0156        & 89.9 & --  & -- \\
Ours                             & \best{0.0083} & 55.2 & \best{3/4} & \best{5/5} & \best{0.0094} & 90.2 & \best{6/6} & \best{2/3} \\
\midrule
one direction                    & 0.0176        & 54.8 & 0/4 & \best{5/5} & 0.0102        & 90.3 & \best{6/6} & 0/3 \\
without anchor                   & 0.0142        & 55.0 & 0/4 & 4/5 & 0.0112        & 90.8 & 5/6 & 1/3 \\
\bottomrule
\end{tabular}
\caption{Ablation under both protocols.
Full version of Table~\ref{tab:ablation}.}
\label{tab:ablation-full}
\par\vspace{1.5\baselineskip}
\centering
\small
\setlength{\tabcolsep}{5pt}
\begin{tabular}{lrrrr}
\toprule
 & & & \multicolumn{2}{c}{Variants of Table~\ref{tab:ablation}} \\
\cmidrule(lr){4-5}
Topic & Unadapted & Ours & one direction & without anchor \\
\midrule
\multicolumn{5}{l}{\textit{In-dataset (RoMQA)}} \\
Actors and Filmmakers & $+$0.0211 & \best{$+$0.0165} & $+$0.0495 & $+$0.0223 \\
Computer science & $-$0.0072 & $+$0.0043 & $+$0.0043 & \best{$-$0.0010} \\
Former countries & $-$0.0121 & \best{$+$0.0041} & \best{$+$0.0041} & $+$0.0163 \\
Human rights & $-$0.0356 & $+$0.0104 & $+$0.0104 & \best{$-$0.0035} \\
Literature & $+$0.0024 & \best{$+$0.0010} & $+$0.0026 & $+$0.0041 \\
Medicine & \best{$+$0.0046} & $+$0.0058 & $+$0.0084 & $+$0.0080 \\
Netherlands & $+$0.0194 & \best{$-$0.0051} & $+$0.0518 & $+$0.0203 \\
Sociology & $-$0.0092 & \best{$-$0.0019} & \best{$-$0.0019} & $-$0.0077 \\
Southeast Asia & $-$0.0409 & \best{$-$0.0255} & \best{$-$0.0255} & $-$0.0452 \\
\cmidrule(lr){2-5}
$|GD|$ & 0.0169 & \best{0.0083} & 0.0176 & 0.0142 \\
Accuracy              & 55.1 & 55.2 & 54.8 & 55.0 \\
\midrule
\multicolumn{5}{l}{\textit{Cross-dataset (TREC)}} \\
Actors and Filmmakers & $+$0.0385 & \best{$+$0.0074} & \best{$+$0.0074} & $+$0.0174 \\
Computer science & \best{$+$0.0002} & $-$0.0005 & $-$0.0005 & $-$0.0006 \\
Former countries & $+$0.0083 & \best{$-$0.0076} & \best{$-$0.0076} & \best{$-$0.0076} \\
Human rights & \best{$-$0.0073} & $-$0.0075 & $-$0.0103 & $-$0.0075 \\
Literature & $-$0.0031 & \best{$-$0.0029} & $-$0.0031 & $-$0.0033 \\
Medicine & $+$0.0343 & \best{$+$0.0226} & \best{$+$0.0226} & \best{$+$0.0226} \\
Netherlands & $-$0.0049 & \best{$+$0.0018} & $-$0.0050 & $+$0.0053 \\
Sociology & $+$0.0396 & \best{$+$0.0326} & \best{$+$0.0326} & \best{$+$0.0326} \\
Southeast Asia & $+$0.0043 & \best{$+$0.0023} & \best{$+$0.0023} & $+$0.0044 \\
\cmidrule(lr){2-5}
$|GD|$ & 0.0156 & \best{0.0094} & 0.0102 & 0.0112 \\
Accuracy              & 89.9 & 90.2 & 90.3 & 90.8 \\
\bottomrule
\end{tabular}
\caption{$GD_t$ of every variant of Table~\ref{tab:ablation}, beside \emph{Unadapted} and \emph{Ours}, under both protocols.
The rows $|GD|$ and Accuracy give, for each column, the mean absolute per-topic group disparity and the accuracy that Table~\ref{tab:ablation-full} reports.
Bold marks, in each row, the value closest to zero.}
\label{tab:ablation-per-topic}
\end{table*}

\end{document}